\documentclass[10pt,a4paper]{article}

\usepackage[T1]{fontenc}
\usepackage[utf8x]{inputenc}

\usepackage[english]{babel}
\usepackage[numbers,sort&compress]{natbib}

\usepackage{anysize}
\marginsize{3cm}{3cm}{3cm}{3cm}

\usepackage{titlesec}
\titlespacing{\section}{0pt}{\parskip}{0pt}
\titlespacing{\subsection}{0pt}{\parskip}{0pt}
\titlespacing{\subsubsection}{0pt}{\parskip}{0pt}

\usepackage{enumitem}
\setlist[itemize]{itemsep=0pt, topsep=0pt}
\setlist[enumerate]{itemsep=0pt, topsep=0pt}

\usepackage[hang,footnotesize,bf]{caption}
\usepackage{subcaption}
\usepackage[usenames, dvipsnames]{color}
\usepackage{graphicx}
\graphicspath{{./images/}}
\DeclareGraphicsExtensions{.jpg}

\usepackage{amsmath}
\usepackage{amsfonts}
\usepackage{amssymb}

\usepackage{url}
\usepackage[hang,flushmargin]{footmisc}
\usepackage{hyperref}
\hypersetup{      
    unicode=true,         
    colorlinks=true,       
    linkcolor=black,
    citecolor=black,
    filecolor=black,
    urlcolor=black
}

\usepackage{tocloft}
\usepackage{float}
\usepackage{listings}
\usepackage{framed}
\usepackage{textcomp}
\usepackage{titlesec}
\usepackage{placeins}

\titleformat{\section}{\normalfont\Large\bfseries}{\thesection}{1em}{}[{\titlerule[0.8pt]}]

\newcommand{\registered}{\raisebox{0.75ex}{\tiny{\textregistered}}}
\makeatletter
\def\blfootnote{\xdef\@thefnmark{}\@footnotetext}
\makeatother

\title{CREATE: Multimodal Dataset for Unsupervised Learning, Generative Modeling and Prediction of Sensory Data from a Mobile Robot in Indoor Environments}
\author{Simon Brodeur, Simon Carrier, Jean Rouat \\ 
\multicolumn{1}{p{.9\textwidth}}{\centering\emph{Département de génie électrique et génie informatique, \\
Université de Sherbrooke, Canada}\\ \vspace{1.5ex} \url{simon.brodeur@usherbrooke.ca}}}
\date{January 2018}

\begin{document}

\maketitle

\paragraph{Abstract}
The CREATE database is composed of 14 hours of multimodal recordings from a mobile robotic platform based on the iRobot Create\registered. The various sensors cover vision, audition, motors and proprioception. The dataset has been designed in the context of a mobile robot that can learn multimodal representations of its environment, thanks to its ability to navigate the environment. This ability can also be used to learn the dependencies and relationships between the different modalities of the robot (e.g. vision, audition), as they reflect both the external environment and the internal state of the robot. The provided multimodal dataset is expected to have multiple usages:
\begin{itemize}
\item Multimodal unsupervised object learning: learn the statistical regularities (structures) of the sensor inputs per modality and across modalities.
\item Multimodal prediction: learn to predict future states of the sensory inputs. 
\item Egomotion detection: learn to predict motor states from the other sensory inputs (e.g. visual optical flow, gyroscope).
\item Causality detection: learn to predict when the robot affects its own sensory inputs (i.e. due to motors), and when the environment is perturbing the sensory inputs (e.g. the user moves the robot around, the robot sees a human moving).
\end{itemize}

\hfill

\blfootnote{Copyright and Trademark Notice: iRobot, Roomba and Create are registered trademarks of iRobot Corporation.}

\clearpage \newpage
\tableofcontents
\newpage

\section{Mobile Robotic Platform}

In this section, the hardware and software components used to record the CREATE dataset are explained.
Aside from the custom CNC-cut clear acrylic mounting plates, the mobile robotic platform is made all of off-the-shelf components.
We thus aimed to use accessible hardware and open-source software. Robotic platforms similar to ours have been used to record datasets for cooperative localization and mapping \citep{Keith2011}, or for speech and audio research \citep{LeRoux2015}. Such realistic datasets help the development of offline learning algorithms.

\subsection{Hardware}

\begin{figure}[tb]
    \centering
    \includegraphics[width=0.95\textwidth]{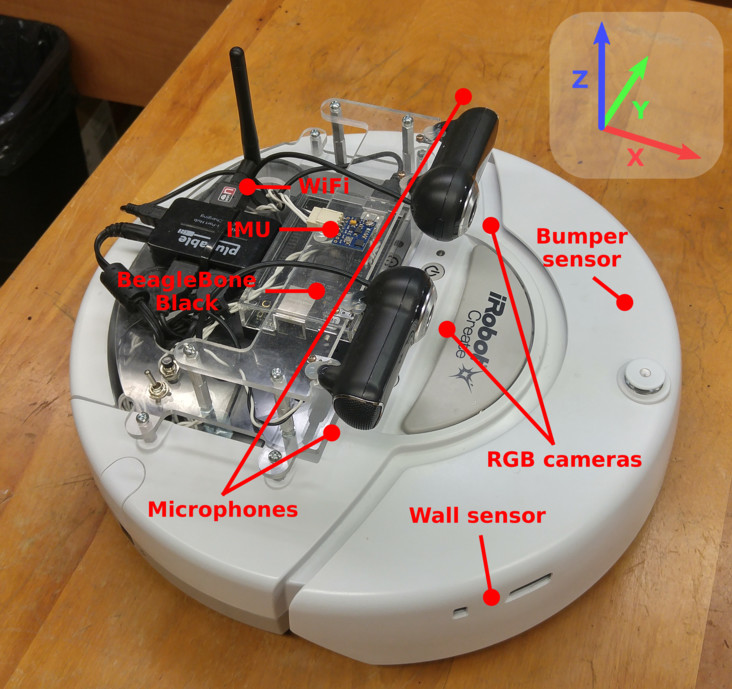}
    \caption{The iRobot Create\registered~ mobile robot used for all experiments. Some of the most important components are shown in red. On the top right of the image, the X-Y-Z coordinate system used is shown.}
    \label{fig:create-robot}
\end{figure}

The iRobot Create\registered~ mobile robot was used because of its low price and ease of access by the scientific and public community. Since its released, this robot has been used as a moving base platform for several robotic projects, and even derived commercial products such as the original TurtleBot \citep{Turtlebot}. In our case, a Beaglebone Black \citep{Beagleboard} embedded computer was added to the robot for control and data acquisition. The Beaglebone Black was running Debian Linux 8.0 \citep{Debian}. The various sensors were connected to the embedded computer using the available serial, i2C and USB ports. The following parts constitute the mobile robotic platform used for the dataset:
\begin{itemize}
\item iRobot Create\registered~ programmable robot\footnote{Note that the original iRobot Create\registered~was discontinued in 2014. The company iRobot has since released the iRobot Create\registered~2 \citep{Create}, which features similar and even improved functionalities.}.
\item BeagleBone Black with acrylic case.
\item Plugable USB 2.0 4-Port Hub with 12.5 W Power Adapter with BC 1.2 Charging.
\item Dual Logitech Quickcam Pro 9000 camera.
\item 5 A DC-DC Adjustable Buck Step Down Module 24V/12V/5V Voltage Regulator Converter (unregulated battery to 5 V).
\item Real Time Clock Memory Module (DS3231) .
\item 10DOF IMU: accelerometer and magnetometer (LSM303D), gyroscope (L3GD20), pressure sensor (BMP180).
\item 802.11n/g/b 150Mbps Mini USB WiFi Wireless Adapter Network LAN Card w/Antenna (Ralink RT5370).
\end{itemize}

For remote control, the following controllers were used:
\begin{itemize}
\item Wired USB Vibration Shock Gamepad Game Controller Joystick Joypad for PC Laptop.
\item Xbox 360 Wireless Controller, with Xbox 360 Wireless Gaming Receiver.
\end{itemize}

\subsection{Software}

The software developed for the robot is based on the Robotic Operating System (ROS)\citep{ROS}.
Whenever possible, standardized messages and topics were used to provide more compatibility with the ROS community. The complete source code for the tools developed is available online with the dataset. Note that because of the low computing power resources of that BeagleBone Black, the acquisition chain and choices of the sampling rates for each modality were carefully designed.
The dataset was then converted from a ROS-specific format (rosbag) to the more popular HDF5 format.

\section{Dataset specifications}

\subsection{Summary}

The following sensors were recorded and made available in the CREATE dataset:
\begin{itemize}
\item Left and right RGB cameras (320x240, JPEG, 30 Hz sampling rate)
\item Left and right optical flow fields (16x12 sparse grid, 30 Hz sampling rate)
\item Left and right microphones (16000 Hz sampling rate, 64 ms frame length)
\item Inertial measurement unit: accelerometer, gyroscope, magnetometer (90 Hz sampling rate)
\item Battery state (50 Hz sampling rate)
\item Left and right motor velocities (50 Hz sampling rate)
\item Infrared and contact sensors (50 Hz sampling rate)
\item Odometry (50 Hz sampling rate)
\item Atmospheric pressure (50 Hz sampling rate)
\item Air temperature (1 Hz sampling rate)
\end{itemize}

Other relevant information about the recordings is also included:
\begin{itemize}
\item Room location, date and time of the session.
\item Stereo calibration parameters for the RGB cameras.
\end{itemize}

\subsection{Dataset File Format}

The data is provided as a set of HDF5 files, one per recording session. The files are named to include the room and session identifiers, as well as the recording date and time (ISO 8601 format). The recording sessions related to a particular experiment are stored in a separate folder. Overall, the file hierarchy is as follows:
\vspace{-2ex}
\begin{center}
\begin{verbatim}
<EXPERIMENT_ID>/<ROOM_ID>/<EXPERIMENT_ID>_<ROOM_ID>_<SESSION_ID>_<DATETIME>.h5
\end{verbatim}
\end{center}

\clearpage
\subsection{HDF5 File Structure}

All sensors were recorded asynchronously and accurate timestamps are provided for all data samples.
The timestamps are stored as 64 bits floating point values representing the elapsed time since the beginning of the recording, in seconds.
In the HDF5 files, the timestamps are stored under the name \textit{clock} for each sensor (e.g. \verb!/video/left/clock!). The associated data is stored under the name \textit{data} for each sensor (e.g. \verb!/video/left/data!). The JPEG-encoded video frames of the left and right cameras have variable lengths, up to about 40,000 bytes. In that case, \textit{data} was allocated for that maximum size, but partially filled for each frame. The actual byte length of each frame is then stored under the name \textit{shape} (e.g. \verb!/video/left/shape!). Because the HDF5 file is compressed with GZIP, the unfilled values in \textit{data} stays to zero and are thus easily compressible.)

\subsubsection{Audio}

Audio was recorded from the Logitech Quickcam Pro 9000 cameras, providing stereo channels at a sampling rate of 16000 Hz and a sample resolution of 16 bit (signed). The distance between the microphones was measured as 200 mm. The audio data is stored in the HDF5 files as per-channel, time-stamped frames of 1024 samples (i.e. 64 ms). The timestamps represent the time at the end of the acquisition of each frame. 
In Experiment I and Experiment III, the sounds recorded mostly represents:
\begin{itemize}
\item The noise made by the motors when running.
\item The vibrations transmitted to the microphones through the chassis.
\item The friction of the wheels on the floor or any bump or change of material encountered on its surface.
\item The collision of the front bumper with obstacles (e.g. wall, chair).
\end{itemize}
In Experiment II, since there were passive humans in the scene, sometime speech was also recorded additionally to all the sounds described above.
Note that the left and right audio channels were recorded asynchronously, thus they may not contain exactly the same number of frames.
This also means that audio localization based on interaural time difference (ITD) may not be possible, because we did not precisely control the acquisition across both channels nor the specific USB latencies.

\begin{minipage}{\textwidth}
\begin{lstlisting}[frame=single, basicstyle=\footnotesize\ttfamily,
caption= {An example HDF5 hierarchy for audio data in the dataset.},captionpos=b]
GROUP "/" {
   GROUP "audio" {
      GROUP "left" {
         DATASET "clock" {
            DATATYPE  H5T_IEEE_F64LE
            DATASPACE  SIMPLE { ( 14007 ) / ( 14007 ) }
         }
         DATASET "data" {
            DATATYPE  H5T_STD_I16LE
            DATASPACE  SIMPLE { ( 14007, 1024 ) / ( 14007, 1024 ) }
         }
      }
      GROUP "right" {
         DATASET "clock" {
            DATATYPE  H5T_IEEE_F64LE
            DATASPACE  SIMPLE { ( 14012 ) / ( 14012 ) }
         }
         DATASET "data" {
            DATATYPE  H5T_STD_I16LE
            DATASPACE  SIMPLE { ( 14012, 1024 ) / ( 14012, 1024 ) }
         }
      }
   }
}
\end{lstlisting}
\end{minipage}

\begin{figure}[!htb]
    \centering
    \begin{subfigure}[b]{0.48\textwidth}
        \includegraphics[width=\textwidth]{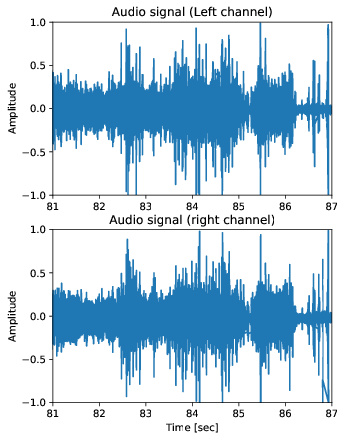}
    \end{subfigure}
     ~ 
    \begin{subfigure}[b]{0.48\textwidth}
        \includegraphics[width=\textwidth]{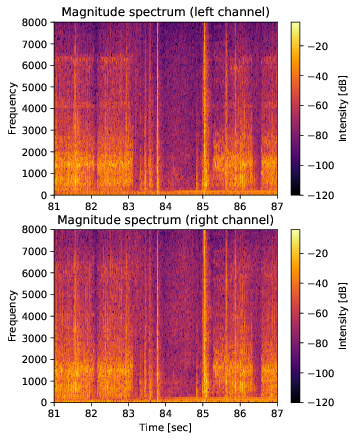}
    \end{subfigure}
    \caption{Example of audio signals (time domain signals and magnitude spectrograms) from the dataset. The magnitude spectrogram was calculated using a window of 256 points, with 50\% overlap. In the illustrated signals, the robot moved forward, changed direction, stopped and accelerated again. The section with no motor activations (when stopped) is visible in the magnitude spectrogram in the interval [84, 85] sec. Overall, the audio signals are very noisy but feature strong temporal and spectral structures related to the motors.}
\end{figure}

\subsubsection{Battery}

The iRobot Create\registered~ used the Advanced Power System (APS) 3000 mAh NiMH battery, with a rated voltage of 14.4 V. At full charge, the voltage could reach almost 16 V. All battery-related signals were sampled at 50 Hz from the iRobot Create\registered~ onboard controller. The battery powered the robot's onboard controller and motors, as well as the additional BeagleBone Black controller, the powered USB hub and all sensors that connected to it (e.g. Logitech Quickcam Pro 9000 cameras and Wi-Fi wireless adapter). This battery provided enough capacity to run the robot for at least an hour, thus enough to record 4 x 15 min sessions on one charge.

\begin{table}[!htb]
\centering
\begin{tabular}{l|c|c|c}
\hline
Signal & Units & Approx. Data Range & Dimension \\ \hline
Voltage & V & [0, 16] & 0 \\
Current & A & [-2, 0] & 1 \\
Charge  & Ah & [0, 3] & 2 \\
Capacity & Ah & [0, 3] & 3 \\
Percentage & \% (as fraction) & [0, 1] & 4 \\
\end{tabular}
\caption{Description of battery-related signals in the dataset. The current drawn from the battery was negative during normal load (i.e. draining), and positive during charging. No charging actually occurred in the dataset.}
\end{table}

\begin{minipage}{\textwidth}
\begin{lstlisting}[frame=single, basicstyle=\footnotesize\ttfamily,
caption= {An example HDF5 hierarchy for battery data in the dataset.},captionpos=b]
GROUP "/" {
   GROUP "battery" {
      GROUP "charge" {
         DATASET "clock" {
            DATATYPE  H5T_IEEE_F64LE
            DATASPACE  SIMPLE { ( 44848 ) / ( 44848 ) }
         }
         DATASET "data" {
            DATATYPE  H5T_IEEE_F64LE
            DATASPACE  SIMPLE { ( 44848, 5 ) / ( 44848, 5 ) }
         }
      }
   }
}
\end{lstlisting}
\end{minipage}

\begin{figure}[!htb]
    \centering
    \begin{subfigure}[b]{1.0\textwidth}
        \includegraphics[width=\textwidth]{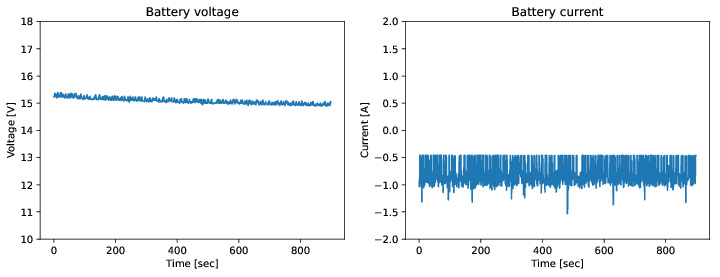}
    \end{subfigure}
    \begin{subfigure}[b]{1.0\textwidth}
        \includegraphics[width=\textwidth]{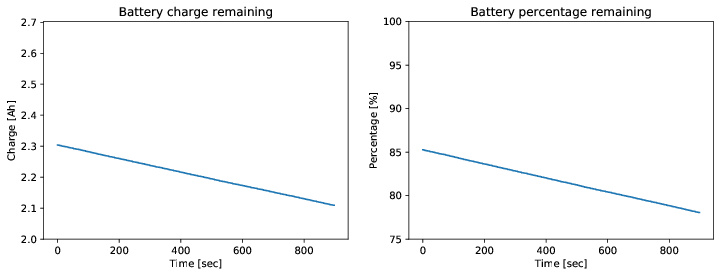}
    \end{subfigure}
    \caption{Example of battery-related signals from the dataset. In the illustrated example, the robot navigated a room for an entire recording session (i.e. 15 min). The battery charge and percentage remaining show a slow and predictable reduction in available charge over time. The battery voltage and current show more fine detail structures related to the activations of the motors, that induced current draw spikes. Note that sometimes, the onboard controller would report a high available charge in the first few seconds of a recording session, and then drop to a more realistic value.}
\end{figure}

\clearpage
\subsubsection{Collision: Contact Switches}

The iRobot Create\registered~ features several contact sensors either based on mechanical sensors (e.g. bumper sensors) or infrared sensors (e.g. wall and cliff sensors).
They are used to detect cliffs on the floor surface, walls on the right side of the robot or when the robot is being lifted from the ground.
The output of those sensors represents binary activations, and are stored as 8 bit unsigned values (i.e. 0 = no contact, 1 = contact).
All collision-related signals were sampled at 50 Hz from the iRobot Create\registered~ onboard controller.

\begin{table}[!htb]
\centering
\begin{tabular}{l|c|c|c}
\hline
Signal & Units & Approx. Data Range & Dimension \\ \hline
Left bumper & binary & [0, 1] & 0 \\
Right bumper & binary & [0, 1] & 1 \\
Caster wheel drop & binary & [0, 1] & 2 \\
Left wheel drop & binary & [0, 1] & 3 \\
Right wheel drop & binary & [0, 1] & 4 \\
Left cliff & binary & [0, 1] & 5 \\
Front-left cliff & binary & [0, 1] & 6 \\
Front-right cliff  & binary & [0, 1] & 7 \\
Right cliff & binary & [0, 1] & 8 \\
Wall & binary & [0, 1] & 9 \\
\end{tabular}
\caption{Description of contact-related sensor signals in the dataset.}
\end{table}

\begin{minipage}{\textwidth}
\begin{lstlisting}[frame=single, basicstyle=\footnotesize\ttfamily,
caption= {An example HDF5 hierarchy for collision data related to the contact switches in the dataset.},captionpos=b]
GROUP "/" {
   GROUP "collision" {
      GROUP "switch" {
         DATASET "clock" {
            DATATYPE  H5T_IEEE_F64LE
            DATASPACE  SIMPLE { ( 44936 ) / ( 44936 ) }
         }
         DATASET "data" {
            DATATYPE  H5T_STD_U8LE
            DATASPACE  SIMPLE { ( 44936, 10 ) / ( 44936, 10 ) }
         }
      }
   }
}
\end{lstlisting}
\end{minipage}

\begin{figure}[!htb]
    \centering
    \begin{subfigure}[b]{1.0\textwidth}
        \includegraphics[width=\textwidth]{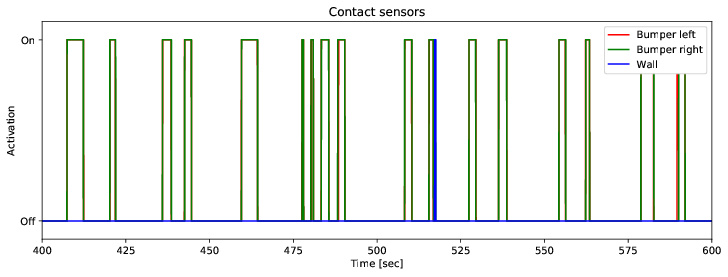}
    \end{subfigure}
    \begin{subfigure}[b]{1.0\textwidth}
        \includegraphics[width=\textwidth]{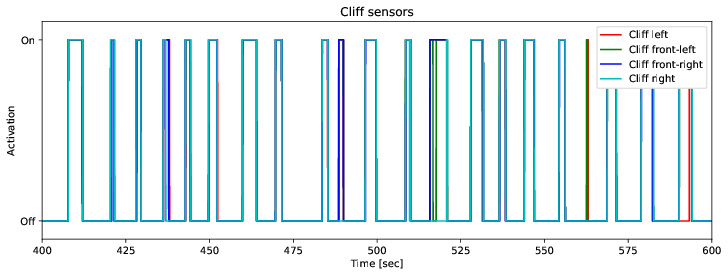}
    \end{subfigure}
    \begin{subfigure}[b]{1.0\textwidth}
        \includegraphics[width=\textwidth]{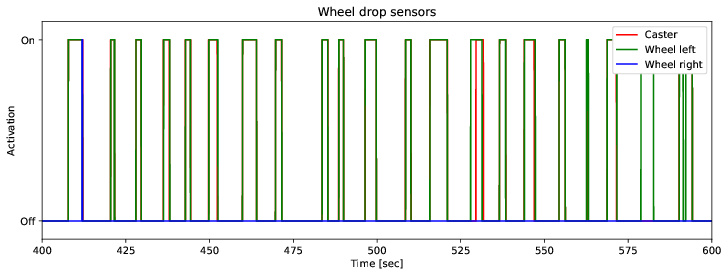}
    \end{subfigure}
    \caption{Example of bumper, cliff and wheel-drop sensor signals from the dataset. In the illustrated example, the robot navigated a room while being frequently manipulated by the human experimenter for a few minutes. The activations visible on the bumper sensors (e.g. at t = 460 sec) are due to the experimenter grabbing the robot by the bumper. The activations visible on the cliff and wheel drop sensors (e.g. at t = 530 sec) are due to the experimenter lifting the robot off the ground.}
\end{figure} 
  
\subsubsection{Collision: Infrared Rangers}

The iRobot Create\registered~ features several infrared sensors under the bumper to detect cliffs in the floor surface.
It also features an infrared sensor on the right side of the bumper to estimate the distance to a wall.
All those infrared sensors are represented with 64 bits floating point values, but were derived from the 16 bits unsigned values returned by the onboard controller. A value close to zero means no infrared light is returned to the sensor, hence there are no obstacles in the line of sight (e.g. wall or floor surface). The output of the infrared sensors also depends on the absorption property of the surface. This means a black carpet on the floor will output lower sensors values that a bright concrete floor. All collision-related signals were sampled at 50 Hz from the iRobot Create\registered~ onboard controller.

\begin{table}[!htb]
\centering
\begin{tabular}{l|c|c|c}
\hline
Signal & Units & Approx. Data Range & Dimension \\ \hline
Wall & binary & [0, 1] & 0 \\
Left cliff & binary & [0, 1] & 1 \\
Front-left cliff & binary & [0, 1] & 2 \\
Front-right cliff  & binary & [0, 1] & 3 \\
Right cliff & binary & [0, 1] & 4 \\
\end{tabular}
\caption{Description of infrared-related signals in the dataset.}
\end{table}

\begin{minipage}{\textwidth}
\begin{lstlisting}[frame=single, basicstyle=\footnotesize\ttfamily,
caption= {An example HDF5 hierarchy for collision data related to the infrared rangers in the dataset.},captionpos=b]
GROUP "/" {
   GROUP "collision" {
      GROUP "range" {
         DATASET "clock" {
            DATATYPE  H5T_IEEE_F64LE
            DATASPACE  SIMPLE { ( 44874 ) / ( 44874 ) }
         }
         DATASET "data" {
            DATATYPE  H5T_IEEE_F64LE
            DATASPACE  SIMPLE { ( 44874, 5 ) / ( 44874, 5 ) }
         }
      }
   }
}
\end{lstlisting}
\end{minipage}

\begin{figure}[!htb]
    \centering
    \includegraphics[width=1.0\textwidth]{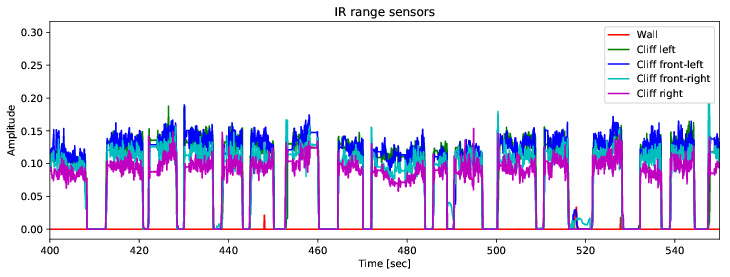}
    \caption{Example of infrared ranger signals from the dataset. In the illustrated example, the robot navigated a room while being frequently manipulated by the human experimenter for a few minutes. When the robot was lifted off the ground by the experimenter (e.g. at t = 410 sec), the amplitude of the cliff sensors dropped to zero. When the robot was in contact or enough near to the ground (e.g. at t = 480 sec), amplitudes vary in the interval [0.05, 0.20] depending on the texture and reflectivity of the floor surface. In that sequence, it can be seen that the wall sensor (shown in red) is rarely active.}
\end{figure}

\subsubsection{Inertial Measurement Unit: Gyroscope, Accelerometer and Magnetometer}

The inertial measurement unit (IMU) attached to the BeagleBone Black included an accelerometer/magnetometer sensor (LSM303D) and a gyroscope (L3GD20) sensor.
The angular velocity (in rad/sec), linear acceleration (in m/sec$^2$) and magnetic field strength (in Tesla) were recorded over all 3 degrees-of-freedom (DOF) of the robot.
Those IMU-related signals were sampled at 90 Hz. No soft-iron or hard-iron correction was performed.
Each recorded at timestamped sample for those sensors is thus a 3-dimensional vector in the X-Y-Z dimension order.
For angular velocity, the L3GD20 sensor was configured for 250 DPS resolution range, thus ranging in the interval [-2.18, 2.18] rad/sec.
For linear acceleration the LSM303D sensor was configured for a 4 G resolution range, thus ranging in the approximated interval [-40, 40] m/sec$^2$.
Note that the we used the X-forward, Y-right and Z-up coordinate system.

\begin{minipage}{\textwidth}
\begin{lstlisting}[frame=single, basicstyle=\footnotesize\ttfamily,
caption= {An example HDF5 hierarchy for angular velocity data in the dataset.},captionpos=b]
GROUP "/" {
   GROUP "imu" {
      GROUP "angular_velocity" {
         DATASET "clock" {
            DATATYPE  H5T_IEEE_F64LE
            DATASPACE  SIMPLE { ( 81312 ) / ( 81312 ) }
         }
         DATASET "data" {
            DATATYPE  H5T_IEEE_F64LE
            DATASPACE  SIMPLE { ( 81312, 3 ) / ( 81312, 3 ) }
         }
      }
   }
}
\end{lstlisting}
\end{minipage}

\begin{minipage}{\textwidth}
\begin{lstlisting}[frame=single, basicstyle=\footnotesize\ttfamily,
caption= {An example HDF5 hierarchy for linear acceleration data in the dataset.},captionpos=b]
GROUP "/" {
   GROUP "imu" {
      GROUP "linear_acceleration" {
         DATASET "clock" {
            DATATYPE  H5T_IEEE_F64LE
            DATASPACE  SIMPLE { ( 81312 ) / ( 81312 ) }
         }
         DATASET "data" {
            DATATYPE  H5T_IEEE_F64LE
            DATASPACE  SIMPLE { ( 81312, 3 ) / ( 81312, 3 ) }
         }
      }
   }
}
\end{lstlisting}
\end{minipage}

\begin{minipage}{\textwidth}
\begin{lstlisting}[frame=single, basicstyle=\footnotesize\ttfamily,
caption= {An example HDF5 hierarchy for magnetic field data in the dataset.},captionpos=b]
GROUP "/" {
   GROUP "imu" {
      GROUP "magnetic_field" {
         DATASET "clock" {
            DATATYPE  H5T_IEEE_F64LE
            DATASPACE  SIMPLE { ( 81152 ) / ( 81152 ) }
         }
         DATASET "data" {
            DATATYPE  H5T_IEEE_F64LE
            DATASPACE  SIMPLE { ( 81152, 3 ) / ( 81152, 3 ) }
         }
      }
   }
}
\end{lstlisting}
\end{minipage}

\begin{figure}[!htb]
    \centering
    \includegraphics[width=1.0\textwidth]{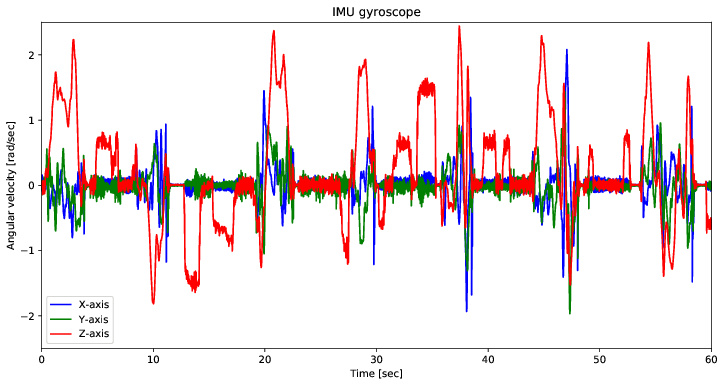}
    \caption{Example of an angular velocity signal from the dataset. In the illustrated example, the robot navigated a room while being frequently manipulated by the human experimenter for a minute. Large angular velocities around the Z-axis (e.g. at t = 14 sec) are often related to the robot changing heading when moving forward. Large angular velocities in the X or Y-axis (e.g. at t = 21 sec) are related to the human manipulating the robot, since when on the ground, the robot can only rotate around the Z-axis.}
    \label{fig:imu-angular-velocity}
\end{figure}

\begin{figure}[!htb]
    \centering
    \includegraphics[width=1.0\textwidth]{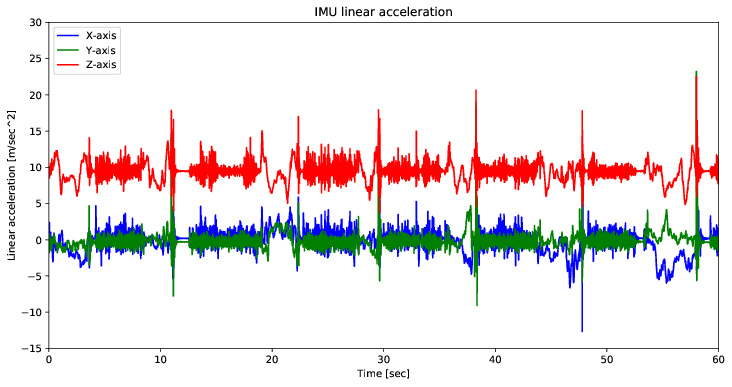}
    \caption{Example of a linear acceleration signal from the dataset. In the illustrated example, the robot navigated a room while being frequently manipulated by the human experimenter for a minute. Note the offset on the Z-axis related to gravity near the typical value of 9.8 $m/s^2$. When the robot is moving forward on the ground (e.g. at t = 40 sec), the vibrations transmitted through the chassis add a visible, high-frequency noise on all axes. When the robot is manipulated by the experimenter (e.g. at t = 20 sec) while being lifted off the ground, the linear accelerations are smoother.}
\end{figure}

\begin{figure}[!htb]
    \centering
    \includegraphics[width=1.0\textwidth]{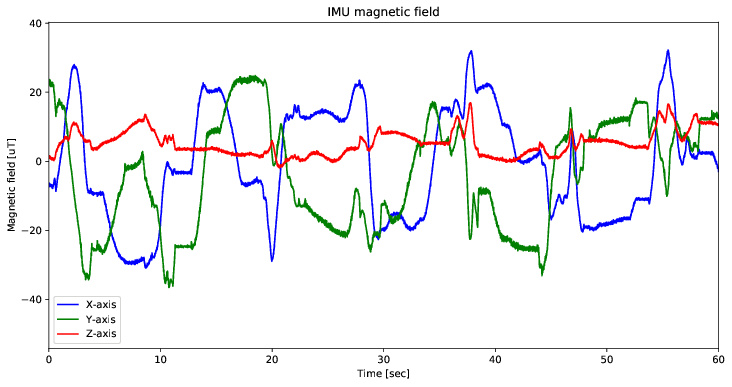}
    \caption{Example of a magnetic field signal from the dataset. In the illustrated example, the robot navigated a room while being frequently manipulated by the human experimenter for a minute. Changes in the magnetic field directions happen smoothly and most often correlates with angular velocity, whether the robot is turning on its own or being moved around by the experimenter. While the magnetometer exhibits less noise than the gyroscope, it is still susceptible in indoor environment to distortions in the magnetic field caused by the steel and concrete skeletons of the building.}
\end{figure}

\clearpage
\subsubsection{Inertial Measurement Unit: Orientation}

The inertial measurement unit (IMU) attached to the BeagleBone Black includes an accelerometer/magnetometer sensor (LSM303D) and a gyroscope (L3GD20) sensor.
The presence of a magnetometer sensor allowed us to get an estimate of the orientation in reference to the north magnetic pole, which can potentially be used for odometry as the absolute orientation in the environment. The standard Madgwick IMU filter from the ROS environment was used to fuse angular velocities, accelerations, and magnetic readings from the IMU into an orientation.
The orientation is estimated at a sampling rate of 90 Hz.
The orientation is represented as a normalized quaternion in the X-Y-Z-W dimension order.

\begin{minipage}{\textwidth}
\begin{lstlisting}[frame=single, basicstyle=\footnotesize\ttfamily,
caption= {An example HDF5 hierarchy for IMU-related orientation data in the dataset.},captionpos=b]
GROUP "/" {
   GROUP "imu" {
      GROUP "orientation" {
         DATASET "clock" {
            DATATYPE  H5T_IEEE_F64LE
            DATASPACE  SIMPLE { ( 80781 ) / ( 80781 ) }
         }
         DATASET "data" {
            DATATYPE  H5T_IEEE_F64LE
            DATASPACE  SIMPLE { ( 80781, 4 ) / ( 80781, 4 ) }
         }
      }
   }
}
\end{lstlisting}
\end{minipage}

\begin{figure}[!htb]
    \centering
    \includegraphics[width=0.70\textwidth]{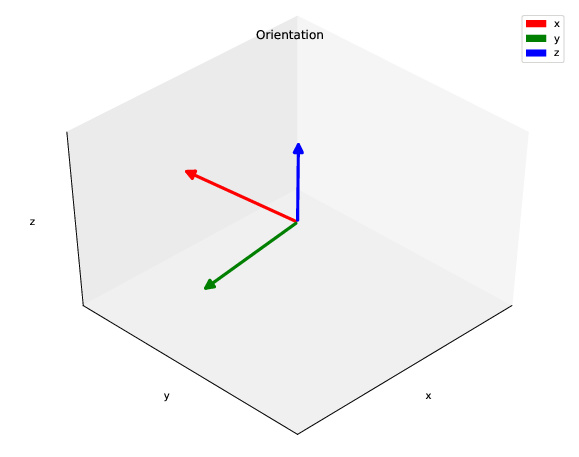}
    \caption{Example of an IMU-related orientation vector from the dataset. Note that the we used the X-forward, Y-right and Z-up coordinate system (illustrated in Figure \ref{fig:create-robot}).}
    \label{fig:imu-orientation}
\end{figure}

\clearpage
\subsubsection{Inertial Measurement Unit: Pressure and Temperature}

The inertial measurement unit (IMU) attached to the BeagleBone Black includes an atmospheric pressure and temperature sensor (BMP180).
The atmospheric pressure is reported in Pascal units, while the temperature is in degree Celsius. While we do not expect those measurements to vary quickly in the recordings,
there could still be small variation of atmospheric pressure recorded when the robot is moving, possibly due to air turbulences caused by the robot itself.
The atmospheric pressure and temperature were respectively sampled at 50 Hz and 1 Hz.
           
\begin{minipage}{\textwidth} 
\begin{lstlisting}[frame=single, basicstyle=\footnotesize\ttfamily,
caption= {An example HDF5 hierarchy for atmospheric pressure and temperature data in the dataset.},captionpos=b]
GROUP "/" {
   GROUP "imu" {
      GROUP "pressure" {
         DATASET "clock" {
            DATATYPE  H5T_IEEE_F64LE
            DATASPACE  SIMPLE { ( 42887 ) / ( 42887 ) }
         }
         DATASET "data" {
            DATATYPE  H5T_IEEE_F64LE
            DATASPACE  SIMPLE { ( 42887, 1 ) / ( 42887, 1 ) }
         }
      }
      GROUP "temperature" {
         DATASET "clock" {
            DATATYPE  H5T_IEEE_F64LE
            DATASPACE  SIMPLE { ( 892 ) / ( 892 ) }
         }
         DATASET "data" {
            DATATYPE  H5T_IEEE_F64LE
            DATASPACE  SIMPLE { ( 892, 1 ) / ( 892, 1 ) }
         }
      }
   }
}
\end{lstlisting}
\end{minipage}

\begin{figure}[!htb]
    \centering
    \begin{subfigure}[b]{0.85\textwidth}
        \includegraphics[width=\textwidth]{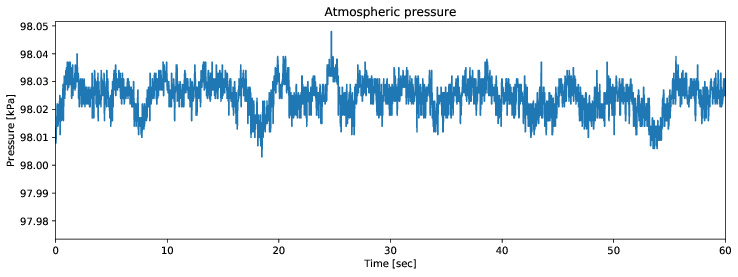}
    \end{subfigure}
    \begin{subfigure}[b]{0.85\textwidth}
        \includegraphics[width=\textwidth]{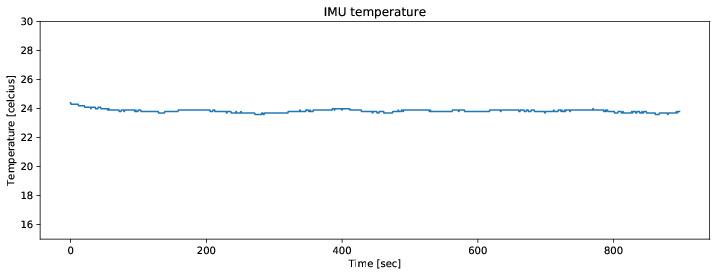}
    \end{subfigure}
    \caption{Example of atmospheric pressure and temperature signals from the dataset. In the illustrated example, the atmospheric pressure is shown for an interval of 60 sec where the robot was navigating a room to show that some fine structure exists, and could possibly be correlated with other sensors. The temperature signal is quite steady over the interval of the entire session (15 min), but still exhibits some variability.}
\end{figure}

\clearpage
\subsubsection{Motors}   

The iRobot Create\registered~ features a differential drive, where the velocity of each wheel (i.e. left and right) can be controlled accurately because of the presence of wheel encoders.
The low-level control with feedback of the wheel encoders was performed by the onboard controller of the iRobot Create\registered~.
All motor-related signals were sampled at 50 Hz from the iRobot Create\registered~ onboard controller.

\begin{table}[!htb]
\centering
\begin{tabular}{l|c|c|c}
\hline
Signal & Units & Approx. Data Range & Dimension \\ \hline
Left wheel & m/sec & [-0.2, 0.2] & 0 \\
Right wheel & m/sec & [-0.2, 0.2] & 1 \\
\end{tabular}
\caption{Description of motor velocity signals in the dataset.}
\end{table}

\begin{minipage}{\textwidth}
\begin{lstlisting}[frame=single, basicstyle=\footnotesize\ttfamily,
caption= {An example HDF5 hierarchy for motor velocity data in the dataset.},captionpos=b]
GROUP "/" {
   GROUP "motor" {
      GROUP "linear_velocity" {
         DATASET "clock" {
            DATATYPE  H5T_IEEE_F64LE
            DATASPACE  SIMPLE { ( 44936 ) / ( 44936 ) }
         }
         DATASET "data" {
            DATATYPE  H5T_IEEE_F64LE
            DATASPACE  SIMPLE { ( 44936, 2 ) / ( 44936, 2 ) }
         }
      }
   }
}
\end{lstlisting}
\end{minipage}

\begin{figure}[!htb]
    \centering
    \includegraphics[width=1.0\textwidth]{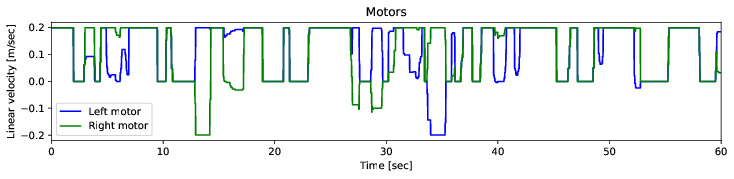}
    \caption{Example of a motor velocity signal from the dataset. In the illustrated example, the robot navigated a room while changing direction. Fast rotations are related to anticorrelated left and right motor velocities (e.g. at t = 14 sec). Forward movements are  correlated with left and right motor velocities (e.g. at t = 25 sec), usually at the maximum allowed velocity of 0.2 m/s. }
    \label{fig:motor-linear-velocity}
\end{figure}

\clearpage
\subsubsection{Odometry: Position and Orientation}   

The inertial measurement unit (IMU) attached to the BeagleBone Black includes an accelerometer/magnetometer sensor (LSM303D) and a gyroscope (L3GD20) sensor and allowed us to estimate the orientation in reference to the north magnetic pole.
The wheel velocities (in m/sec) reported by the onboard iRobot Create\registered~ sensors allowed us to estimate the forward (X-axis) velocity of the robot. Using information about absolute orientation and forward velocity of the robot, the motion was integrated over time in the X-Y-Z axes.
Note that because we could not accurately track motion in the Y-axis (i.e. lateral motion) and Z-axis (i.e. vertical motion), the reported odometry position is restricted to the X-Y plane only. The odometry position over the Z-axis is always zero.
Position is reported over all 3 degrees-of-freedom (DOF), in the X-Y-Z dimension order.

The angular velocities (in rad/sec) are taken from the gyroscope sensor, since a bug in the onboard firmware of the iRobot Create\registered~ didn't allow to pull fast enough the rotation angle estimated from the wheel encoders. 
The linear velocities (in m/sec) used the velocities reported for each wheel to calculate the forward motion in the X-axis. Note that velocities along the Y-axis and Z-axis are always zeros, since the robot can only move along the X-axis or perform in-place rotations.
Both angular and linear velocities are reported over all 3 degrees-of-freedom (DOF), in the X-Y-Z dimension order.

The orientation is represented as a normalized quaternion in the X-Y-Z-W dimension order.
All odometry-related data are estimated at a sampling rate of 50 Hz.

\begin{minipage}{\textwidth}   
\begin{lstlisting}[frame=single, basicstyle=\footnotesize\ttfamily,
caption= {An example HDF5 hierarchy for odometry-related data in the dataset.},captionpos=b]
GROUP "/" {
   GROUP "odometry" {
      GROUP "angular_velocity" {
         DATASET "clock" {
            DATATYPE  H5T_IEEE_F64LE
            DATASPACE  SIMPLE { ( 44845 ) / ( 44845 ) }
         }
         DATASET "data" {
            DATATYPE  H5T_IEEE_F64LE
            DATASPACE  SIMPLE { ( 44845, 3 ) / ( 44845, 3 ) }
         }
      }
      GROUP "linear_velocity" {
         DATASET "clock" {
            DATATYPE  H5T_IEEE_F64LE
            DATASPACE  SIMPLE { ( 44845 ) / ( 44845 ) }
         }
         DATASET "data" {
            DATATYPE  H5T_IEEE_F64LE
            DATASPACE  SIMPLE { ( 44845, 3 ) / ( 44845, 3 ) }
         }
      }
      GROUP "orientation" {
         DATASET "clock" {
            DATATYPE  H5T_IEEE_F64LE
            DATASPACE  SIMPLE { ( 44845 ) / ( 44845 ) }
         }
         DATASET "data" {
            DATATYPE  H5T_IEEE_F64LE
            DATASPACE  SIMPLE { ( 44845, 4 ) / ( 44845, 4 ) }
         }
      }
      GROUP "position" {
         DATASET "clock" {
            DATATYPE  H5T_IEEE_F64LE
            DATASPACE  SIMPLE { ( 44845 ) / ( 44845 ) }
         }
         DATASET "data" {
            DATATYPE  H5T_IEEE_F64LE
            DATASPACE  SIMPLE { ( 44845, 3 ) / ( 44845, 3 ) }
         }
      }
   }
}
\end{lstlisting}
\end{minipage}

\begin{figure}[!htb]
    \centering
    \includegraphics[width=0.70\textwidth]{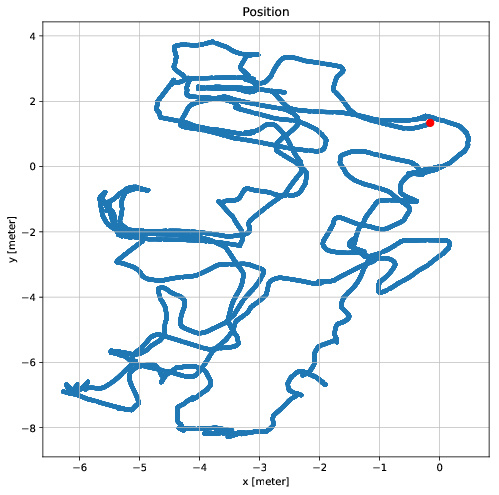}
    \caption{Example of an odometry position 2D trajectory from the dataset. In the illustrated example, the robot navigated a room for an entire recording session (i.e. 15 min). The red point shows its current position. Note that odometry is based on the wheel encoders and IMU-related orientation data, so it strongly suffers from the common drift problem that causes the odometry error to grow over time (e.g. due to wheel slipping or noisy orientation measurements).}
\end{figure}

\begin{figure}[!htb]
    \centering
    \includegraphics[width=0.5\textwidth]{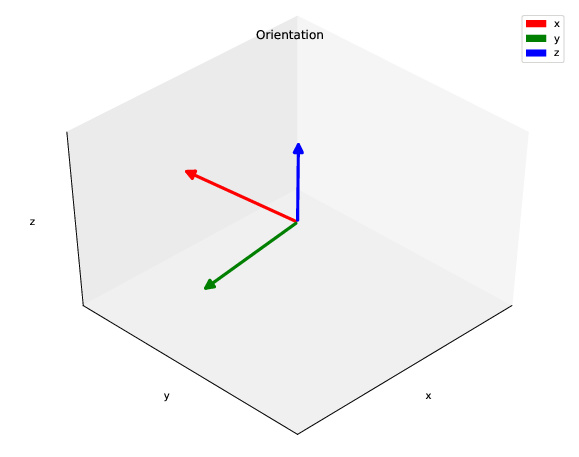}
    \caption{Example of an odometry-related orientation vector from the dataset. Note that the we used the X-forward, Y-right and Z-up coordinate system  (illustrated in Figure \ref{fig:create-robot}). Odometry-related orientation data is almost identical to the IMU-related orientation data (as shown in Figure \ref{fig:imu-orientation}), except that it is sampled at 50 Hz and not 90 Hz. This is because the wheel encoders could not be used to estimate the heading direction.}
\end{figure}

\begin{figure}[htb]
    \centering
    \begin{subfigure}[b]{1.0\textwidth}
        \includegraphics[width=\textwidth]{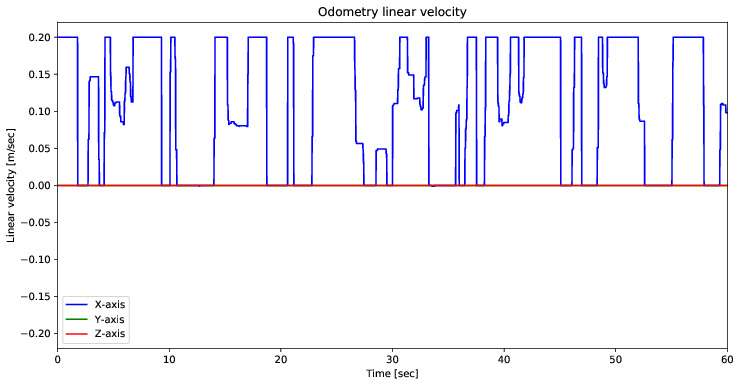}
    \end{subfigure}
    \begin{subfigure}[b]{1.0\textwidth}
        \includegraphics[width=\textwidth]{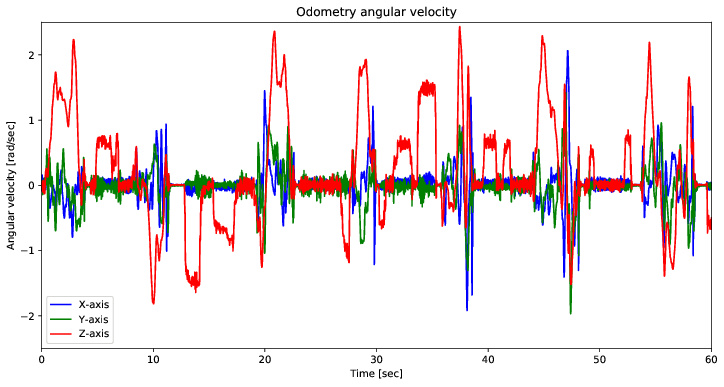}
    \end{subfigure}
    \caption{Example of odometry linear and angular velocity signals from the dataset.  In the illustrated example, the robot navigated a room for a minute. Odometry-related angular velocity data is almost identical to the IMU-related angular velocity data (as shown in Figure \ref{fig:imu-angular-velocity}), except that it is sampled at 50 Hz and not 90 Hz. The odometry-related linear velocity is calculated from the left and right motors (as shown in Figure \ref{fig:motor-linear-velocity}), and is non-zero only on the X-axis (forward axis) because of the 1 DOF constraint of the differential drive. In the sequence above, it can be seen that the robot never went backwards, as the linear velocities stayed positive.}
\end{figure}   
   
\subsubsection{RGB Cameras}   

The stereo camera system is based on two Logitech Quickcam Pro 9000 USB cameras. The cameras were manually adjusted to provide equal contrast and color balance.
The distance between the cameras was measured as 74 mm. 
The image acquisition was done in MJPEG format, at a resolution of 320x240 pixels and a frame rate of 30 Hz.
The raw compressed MJPEG data stream was stored, meaning the low-power BeagleBone Black needed not to decode and re-encode the images during recordings.
Higher frame rates or resolutions could not be achieved without saturating the BeagleBone Black CPU and losing frames during recordings.
Note that while the right camera is physically inverted to properly position the microphone, the image has been inverted in software without needing to be decoded and recompressed.

\begin{minipage}{\textwidth}
\begin{lstlisting}[frame=single, basicstyle=\footnotesize\ttfamily,
caption= {An example HDF5 hierarchy for video data in the dataset.},captionpos=b]
GROUP "/" {
   GROUP "video" {
      GROUP "left" {
         DATASET "clock" {
            DATATYPE  H5T_IEEE_F64LE
            DATASPACE  SIMPLE { ( 26990 ) / ( 26990 ) }
         }
         DATASET "data" {
            DATATYPE  H5T_STD_U8LE
            DATASPACE  SIMPLE { ( 26990, 40000 ) / ( 26990, 40000 ) }
         }
         DATASET "shape" {
            DATATYPE  H5T_STD_I64LE
            DATASPACE  SIMPLE { ( 26990, 1 ) / ( 26990, 1 ) }
         }
      }
      GROUP "right" {
         DATASET "clock" {
            DATATYPE  H5T_IEEE_F64LE
            DATASPACE  SIMPLE { ( 26969 ) / ( 26969 ) }
         }
         DATASET "data" {
            DATATYPE  H5T_STD_U8LE
            DATASPACE  SIMPLE { ( 26969, 40000 ) / ( 26969, 40000 ) }
         }
         DATASET "shape" {
            DATATYPE  H5T_STD_I64LE
            DATASPACE  SIMPLE { ( 26969, 1 ) / ( 26969, 1 ) }
         }
      }
   }
}
\end{lstlisting}
\end{minipage}

\begin{figure}[!htb]
    \centering
    \includegraphics[width=1.0\textwidth]{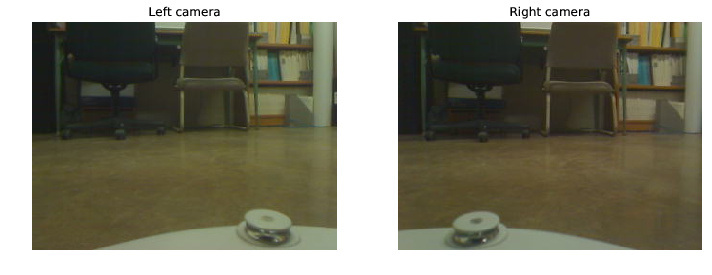}
    \caption{Example of stereo (left and right) camera images from the dataset. Note that a small part of the bumper (including the top bumper IR sensor) is always visible at the bottom right of the visual field.}
\end{figure}

\subsubsection{Optical Flow}

\begin{figure}[!htb]
    \centering
    \includegraphics[width=0.5\textwidth]{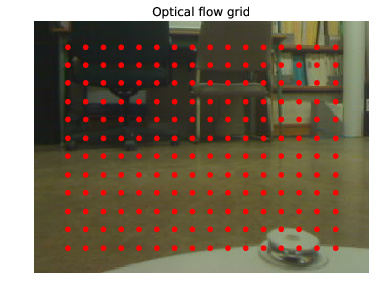}
    \caption{Grid (16 x 12) used to sample the dense optical flow.}
    \label{fig:optical-flow-grid}
\end{figure}

The optical flow was computed using the Gunnar Farneback's algorithm implemented in the OpenCV library \citep{OpenCV}.
The images from the RGB cameras were first converted to greyscale, and a contrast-limiting adaptive histogram equalization (CLAVE) was applied with a clip limit value of 2 and a 8x8 tile grid size. The preprocessed images were then downsampled by a factor of 4.
The optical flow was then calculated for all pixels using pairs of successive images. To increase the sensibility to small variations in optical flow, 
the current frame at index $n$ in the stream was compared with the frame at index $n-5$.
Next, each optical flow field was filtered with a 5 x 5 mean filter to reduce spatial noise.
A grid of 16 x 12 points uniformly spaced (see Figure \ref{fig:optical-flow-grid}) was then sampled from the smoothed optical flow, considering a border of 10\% around the edges of the image. Using a larger grid would have increased significantly the file size of the dataset.
The optical flow was scaled to be in units of pixels/sec relative to the full-size image (i.e. 320x240 pixels).

\begin{minipage}{\textwidth}
\begin{lstlisting}[frame=single, basicstyle=\footnotesize\ttfamily,
caption= {An example HDF5 hierarchy for optical flow data in the dataset.},captionpos=b]
GROUP "/" {
   GROUP "optical_flow" {
      GROUP "left" {
         DATASET "clock" {
            DATATYPE  H5T_IEEE_F64LE
            DATASPACE  SIMPLE { ( 26985 ) / ( 26985 ) }
         }
         DATASET "data" {
            DATATYPE  H5T_IEEE_F64LE
            DATASPACE  SIMPLE { ( 26985, 12, 16, 2 ) / ( 26985, 12, 16, 2 ) }
         }
      }
      GROUP "right" {
         DATASET "clock" {
            DATATYPE  H5T_IEEE_F64LE
            DATASPACE  SIMPLE { ( 26964 ) / ( 26964 ) }
         }
         DATASET "data" {
            DATATYPE  H5T_IEEE_F64LE
            DATASPACE  SIMPLE { ( 26964, 12, 16, 2 ) / ( 26964, 12, 16, 2 ) }
         }
      }
   }
}
\end{lstlisting}
\end{minipage}

\begin{figure}[!htb]
    \centering
    \includegraphics[width=1.0\textwidth]{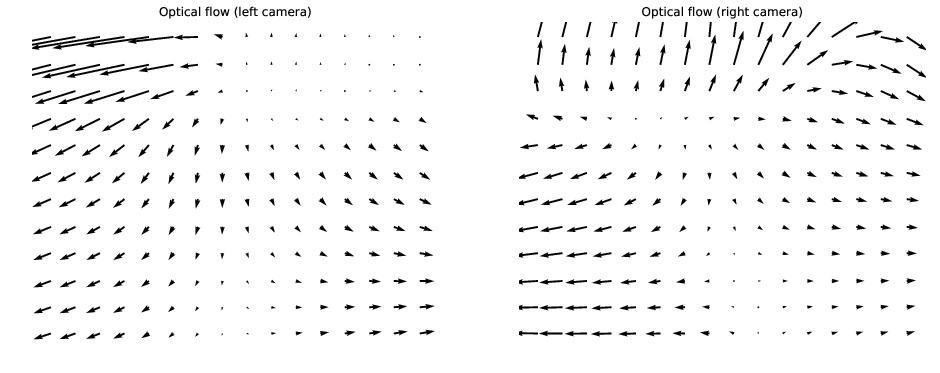}
    \caption{Example of stereo (left and right) optical flow fields from the dataset. The optical flow is illustrated here for each position of the 16x12 grid as the predicted motion of the pixels around each point for the next second (i.e. in units of pixel/sec). Optical flow vectors with large magnitude (e.g. such as the top-left region of the left camera) indicates faster local motion, and the direction of each vector indicates the local motion direction.}
\end{figure}

\section{Experiments}

\subsection{Experiment I: Navigation in Passive Environments}
The robot was moving around a room, controlled by the experimenter using a joystick. Each recorded session was approximately 15 min. There are 4 session recordings per room, with various starting points and trajectories. There was little to no moving objects (including humans) in the room. The robot was directed by the experimenter not to hit any obstacles. 

\begin{table}[!htb]
\centering
\begin{tabular}{ c || c | c | c }
\hline
\textbf{Session} &  \textbf{Truncation} & \textbf{Battery charge} & \textbf{Collision} \\
\hline
\verb!E1_C13035_S1_20161103T175600! &&&x \\
\verb!E1_C13035_S2_20161103T181700! &&&x \\
\verb!E1_C13035_S3_20161103T183500! &&&x \\
\verb!E1_C13035_S4_20161103T185200! &&& \\
\hline
\verb!E1_BTUN_S1_20161106T154732!	&&x&x \\
\verb!E1_BTUN_S2_20161106T160524!	&&&x \\
\verb!E1_BTUN_S3_20161106T162256! &&& \\
\verb!E1_BTUN_S4_20161106T164018! &&& \\ 
\hline
\verb!E1_FLOOR3_S1_20161026T181035! &&&x \\
\verb!E1_FLOOR3_S2_20161028T175439! &&& \\
\verb!E1_FLOOR3_S3_20161028T181205! &&& \\
\verb!E1_FLOOR3_S4_20161101T180521! &x&& \\
\hline
\verb!E1_PROF_S1_20161109T172234! &&x&x \\
\verb!E1_PROF_S2_20161109T175252! &&& \\
\verb!E1_PROF_S3_20161109T181351! &&& \\
\verb!E1_PROF_S4_20161109T183323! &x&& \\
\hline
\verb!E1_C13036_S1_20161127T150642! &&x& \\
\verb!E1_C13036_S2_20161127T152718! &&&x \\
\verb!E1_C13036_S3_20161127T154858! &&&x \\
\verb!E1_C13036_S4_20161127T160955! &x&&x \\
\hline
\verb!E1_C1TUN_S1_20161031T163256! &&& \\
\verb!E1_C1TUN_S2_20161031T165013! &&& \\
\verb!E1_C1TUN_S3_20161031T170718! &&& \\
\verb!E1_C1TUN_S4_20161031T172448! &&& \\
\hline
\verb!E1_C14113_S1_20161102T173800! &&x&x \\	
\verb!E1_C14113_S2_20161102T175600! &&&x \\
\verb!E1_C14113_S3_20161102T181200! &&& \\
\verb!E1_C14113_S4_20161102T183000! &&&x \\
\hline
\verb!E1_C14016_S1_20161027T175408! &&x& \\
\verb!E1_C14016_S2_20161027T181114! &&&x \\
\verb!E1_C14016_S3_20161027T182857! &&& \\
\verb!E1_C14016_S4_20161027T184609! &&& \\
\hline
\end{tabular}
\caption{List of recording sessions for Experiment I. The \textit{truncation} column indicates sessions that needed to be truncated at the end, because too many sensor measurements were lost due to overloaded CPU and memory buffering on the BeagleBone Black. This means those sessions have a duration slightly less than 15 min. The \textit{battery charge} column indicates sessions where the initial remaining battery charge was overestimated by the onboard controller. The \textit{collision} column indicates unintentional collisions as witnessed by the activation of one or more bumper sensors.}
\end{table}

\begin{figure}[!htb]
    \centering
    \begin{subfigure}[b]{0.45\textwidth}
        \includegraphics[width=\textwidth]{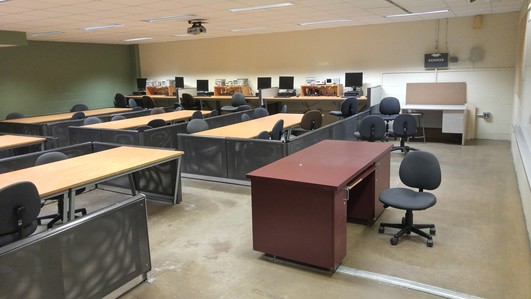} 
        \caption{C1-3035}
    \end{subfigure}
    ~ 
    \begin{subfigure}[b]{0.45\textwidth}
        \includegraphics[width=\textwidth]{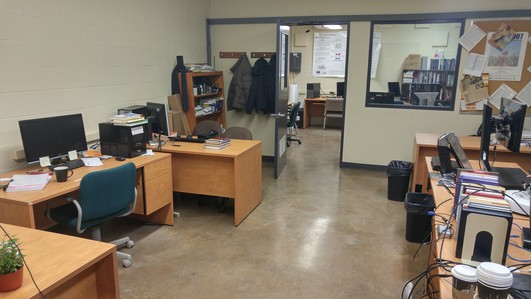} 
        \caption{C1-3036}
    \end{subfigure}
    ~ 
    \begin{subfigure}[b]{0.45\textwidth}
        \includegraphics[width=\textwidth]{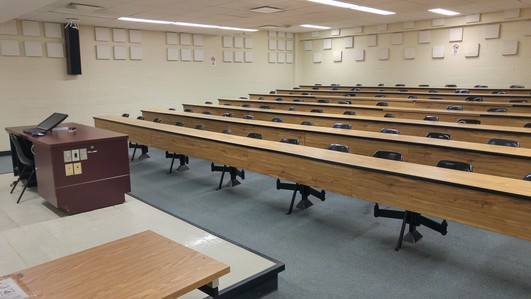} 
        \caption{C1-4016}
    \end{subfigure}
    ~ 
    \begin{subfigure}[b]{0.45\textwidth}
        \includegraphics[width=\textwidth]{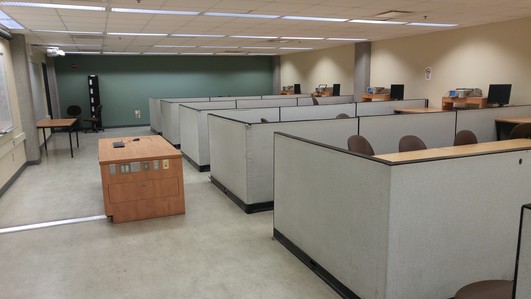} 
        \caption{C1-4113}
    \end{subfigure}
    ~ 
    \begin{subfigure}[b]{0.45\textwidth}
        \includegraphics[width=\textwidth]{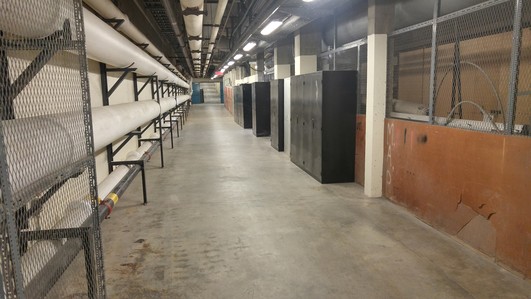} 
        \caption{C1 TUNNEL}
    \end{subfigure}
    ~ 
    \begin{subfigure}[b]{0.45\textwidth}
        \includegraphics[width=\textwidth]{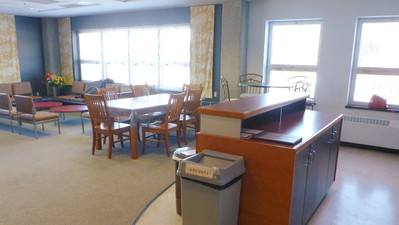} 
        \caption{PROF}
    \end{subfigure}\\
	~ 
    \begin{subfigure}[b]{0.25\textwidth}
        \includegraphics[width=\textwidth]{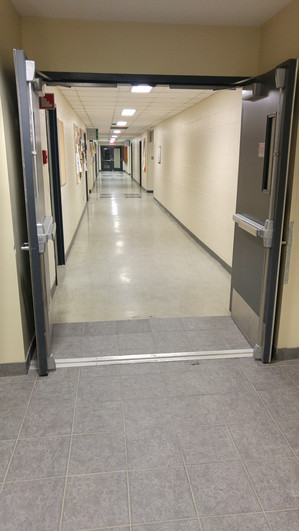} 
        \caption{C1 FLOOR3}
    \end{subfigure}
    ~ 
    \begin{subfigure}[b]{0.25\textwidth}
        \includegraphics[width=\textwidth]{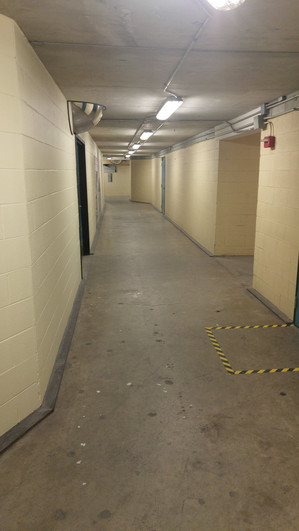} 
        \caption{B TUNNEL}
    \end{subfigure}
    \caption{The various rooms used for Experiment I. They are all locate at Université de Sherbrooke, in the Faculty of Engineering.}
\end{figure}

\subsection{Experiment II: Navigation in Environments with Passive Human Interactions}
In this experiment, the robot was moving around a room, controlled by the experimenter using a joystick. Each recorded session was approximately 15 min. There are 4 session recordings per room, with various starting points and trajectories. Note that compared to Experiment I, there was a significant amount of moving objects (including humans) in the selected rooms.

\begin{table}[!htb]
\centering
\begin{tabular}{ c || c | c | c }
\hline
\textbf{Session} &  \textbf{Truncation} & \textbf{Battery charge} \\
\hline
\verb!E2_C1CORR_S1_20161117T165404! &&x \\
\verb!E2_C1CORR_S2_20161117T171351! &x& \\
\verb!E2_C1CORR_S3_20161117T173323! &x& \\
\verb!E2_C1CORR_S4_20161117T175914! &x& \\
\hline
\verb!E2_FLOOR3_S1_20161201T084316! &&x \\
\verb!E2_FLOOR3_S2_20161208T160537! &&x \\
\verb!E2_FLOOR3_S3_20161208T162946! && \\
\verb!E2_FLOOR3_S4_20161213T162929! &&x \\ 
\hline
\verb!E2_CAFTUN_S1_20161129T121525! && \\
\verb!E2_CAFTUN_S2_20161129T125725! && \\
\verb!E2_CAFTUN_S3_20161129T131337! && \\
\verb!E2_CAFTUN_S4_20161130T115606! && \\
\hline
\verb!E2_B5A1TUN_S1_20161206T121437! &&x \\
\verb!E2_B5A1TUN_S2_20161206T123126! && \\
\verb!E2_B5A1TUN_S3_20161206T124832! && \\
\verb!E2_B5A1TUN_S4_20161206T130506! && \\
\hline
\verb!E2_C1B4TUN_S1_20161208T171018! &&x \\
\verb!E2_C1B4TUN_S2_20161208T172724! && \\
\verb!E2_C1B4TUN_S3_20161208T174332! && \\
\verb!E2_C1B4TUN_S4_20161208T180653! && \\
\hline
\end{tabular}
\caption{List of recording sessions for Experiment II. The \textit{truncation} column indicates sessions that needed to be truncated at the end, because too many sensor measurements were lost due to overloaded CPU and memory buffering on the BeagleBone Black. This means those sessions have a duration slightly less than 15 min. The \textit{battery charge} column indicates sessions where the initial remaining battery charge was overestimated by the onboard controller. No unintentional collision happened during Experiment II.}
\end{table}

\begin{figure}[!htb]
    \centering
    \begin{subfigure}[b]{0.48\textwidth}
        \includegraphics[width=\textwidth]{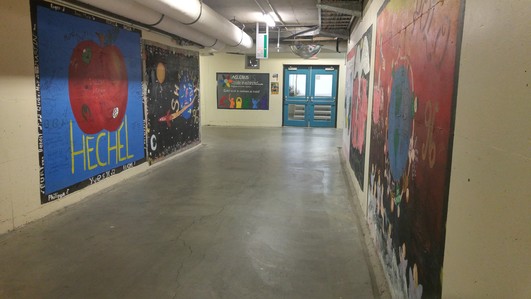} 
        \caption{B5-A1 Tunnels}
    \end{subfigure}
    ~ 
    \begin{subfigure}[b]{0.48\textwidth}
        \includegraphics[width=\textwidth]{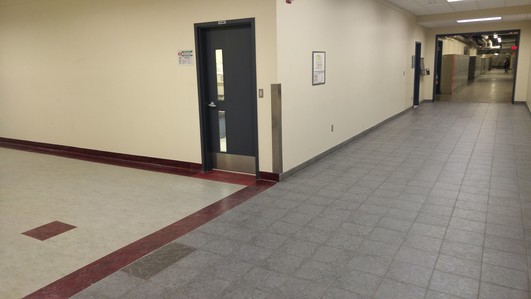} 
        \caption{C1 Corridor}
    \end{subfigure}\\
    \begin{subfigure}[b]{0.30\textwidth}
        \includegraphics[width=\textwidth]{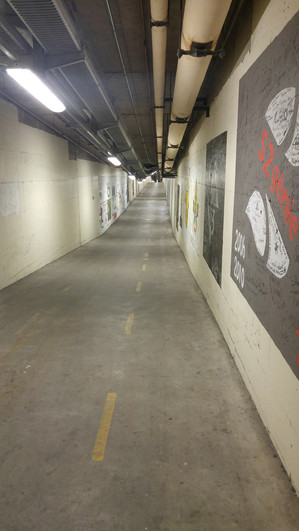} 
        \caption{C1-B4 Tunnels}
    \end{subfigure}
    ~ 
    \begin{subfigure}[b]{0.30\textwidth}
        \includegraphics[width=\textwidth]{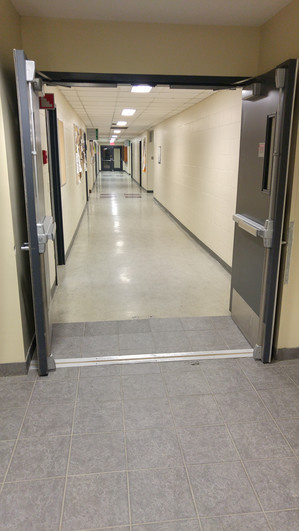} 
        \caption{C1 FLOOR3}
    \end{subfigure}\\
    \begin{subfigure}[b]{0.48\textwidth}
        \includegraphics[width=\textwidth]{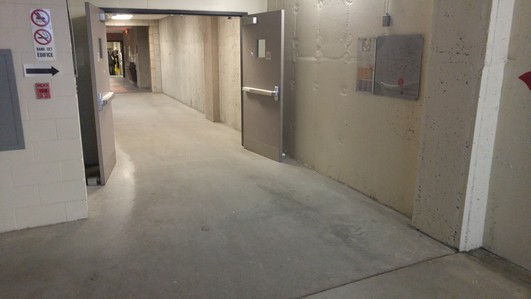} 
        \caption{CAF Tunnels (main cafeteria tunnel)}
    \end{subfigure}
    \caption{The various rooms used for Experiment II. They are all locate at Université de Sherbrooke, in the Faculty of Engineering}
\end{figure}

\subsection{Experiment III: Navigation in Environments with Active Human Interactions}
The robot was moving around a room, controlled by the experimenter using a joystick. A second experimenter lifted the robot and changed its position and orientation at random intervals (e.g. once every 10 sec). Each recorded session was approximately 15 min. There are 5 session recordings in a single room. 

\begin{table}[!htb]
\centering
\begin{tabular}{ c || c | c | c }
\hline
\textbf{Session} &  \textbf{Truncation} \\
\hline
\verb!E3_C13036_S1_20161222T101923! & \\
\verb!E3_C13036_S2_20161222T104206! & \\
\verb!E3_C13036_S3_20161222T113137! & \\
\verb!E3_C13036_S4_20161222T134644! &x \\
\verb!E3_C13036_S5_20161222T140912! & \\
\hline
\end{tabular}
\caption{List of recording sessions for Experiment III. The \textit{truncation} column indicates sessions that needed to be truncated at the end, because too many sensor measurements were lost due to overloaded CPU and memory buffering on the BeagleBone Black. This means those sessions have a duration slightly less than 15 min. No unintentional collision or overestimated battery charge problem happened during Experiment III.}
\end{table}

\begin{figure}[!htb]
    \centering
    \begin{subfigure}[b]{0.48\textwidth}
        \includegraphics[width=\textwidth]{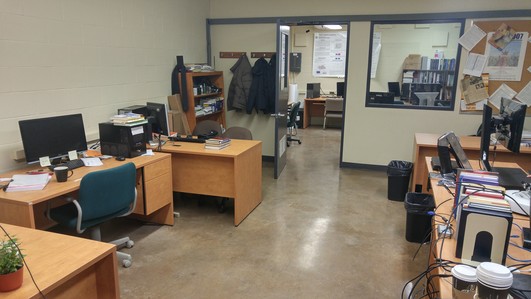} 
    \end{subfigure}
     ~ 
    \begin{subfigure}[b]{0.48\textwidth}
        \includegraphics[width=\textwidth]{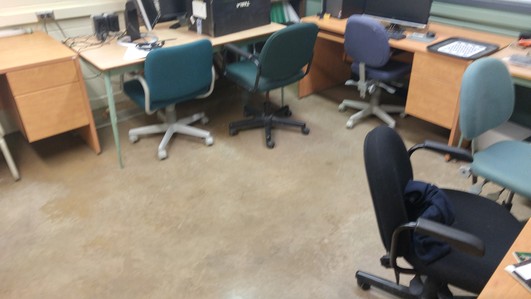} 
    \end{subfigure}
    \caption{The room C1-3036 used for Experiment III.}
\end{figure}

\section{Acknowledgements}

The authors would like to thank the ERA-NET (CHIST-ERA) and FRQNT organizations for funding this research as part of the European IGLU project.

\clearpage
\newpage
\bibliographystyle{ieeetr}
\bibliography{references}

\begin{thebibliography}{1}

\bibitem{Keith2011}
K.~Y. Leung, Y.~Halpern, T.~D. Barfoot, and H.~H. Liu, ``The utias multi-robot
  cooperative localization and mapping dataset,'' {\em The International
  Journal of Robotics Research}, vol.~30, no.~8, pp.~969--974, 2011.

\bibitem{LeRoux2015}
J.~L. Roux, E.~Vincent, J.~R. Hershey, and D.~P.~W. Ellis, ``Micbots:
  Collecting large realistic datasets for speech and audio research using
  mobile robots,'' in {\em 2015 IEEE International Conference on Acoustics,
  Speech and Signal Processing (ICASSP)}, pp.~5635--5639, April 2015.

\bibitem{Turtlebot}
``{TurtleBot} personal robot.'' \url{http://www.turtlebot.com/}.
\newblock Accessed: 2018-01-27.

\bibitem{Beagleboard}
``{BeagleBone Black} open-hardware development platform.''
  \url{https://beagleboard.org/}.
\newblock Accessed: 2018-01-27.

\bibitem{Debian}
``{Debian Linux 8.0 (Jessie)}, a free operating system {(OS)}.''
  \url{https://www.debian.org/}.
\newblock Accessed: 2018-01-27.

\bibitem{Create}
``{iRobot Create 2 Programmable Robot}.'' \url{http://www.irobot.com/}.
\newblock Accessed: 2018-01-27.

\bibitem{ROS}
``{Robot Operating System (ROS)}.'' \url{http://www.ros.org/}.
\newblock Accessed: 2018-01-27.

\bibitem{OpenCV}
``{Open Source Computer Vision Library (OpenCV)}.'' \url{https://opencv.org/}.
\newblock Accessed: 2018-01-27.

\end{thebibliography}

\end{document}